\pdfoutput=1

\documentclass[11pt]{article}

\usepackage{acl}

\usepackage{times}
\usepackage{latexsym}

\usepackage[T1]{fontenc}

\usepackage[utf8]{inputenc}

\usepackage{microtype}
\usepackage{amsmath}
\usepackage{graphicx}
\usepackage{multirow}
\usepackage{mathtools}
\usepackage{makecell}

\title{Unsupervised Abstractive Dialogue Summarization with \\ Word Graphs and POV Conversion}

\author{Seongmin Park, Jihwa Lee\\
  ActionPower, Seoul, Republic of Korea\\
  \texttt{\{seongmin.park, jihwa.lee\}@actionopwer.kr}}

\begin{document}
\maketitle
\begin{abstract}
We advance the state-of-the-art in unsupervised abstractive dialogue summarization by utilizing multi-sentence compression graphs. Starting from well-founded assumptions about word graphs, we present simple but reliable path-reranking and topic segmentation schemes. Robustness of our method is demonstrated on datasets across multiple domains, including meetings, interviews, movie scripts, and day-to-day conversations. We also identify possible avenues to augment our heuristic-based system with deep learning. We open-source our code\footnote{https://github.com/seongminp/graph-dialogue-summary}, to provide a strong, reproducible baseline for future research into unsupervised dialogue summarization. 
\end{abstract}

\section{Introduction}

Compared to traditional text summarization, dialogue summarization introduces a unique challenge: conversion of first- and second-person speech into third-person reported speech. Such discrepancy between the observed text and expected model output puts greater emphasis on abstractive transduction than in traditional summarization tasks. The disorientation is further exacerbated by each of many diverse dialogue types calling for a differing form of transduction – short dialogues require terse abstractions, while meeting transcripts require summaries by agenda. 

Thus, despite the steady emergence of dialogue summarization datasets, the field of dialogue summarization is still bottlenecked by a scarcity of training data. To train a truly robust dialogue summarization model, one requires transcript-summary pairs not only across diverse \textit{dialogue domains}, but also across multiple \textit{dialogue types} as well. The lack of diverse annotated summarization data is especially pronounced in low-resourced languages. From such state of the literature, we identify a need for unsupervised dialogue summarization.  

\begin{figure}[t]
  \centering
  \includegraphics[width=\linewidth]{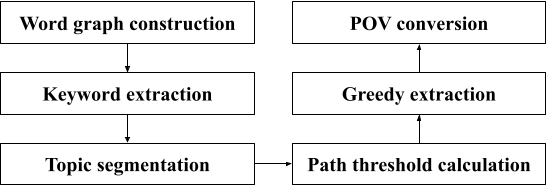}
  \caption{Our summarization pipeline.}
  \label{fig:pipeline}
\end{figure}

Our method builds upon previous research on unsupervised summarization using word graphs. Starting from the simple assumption that \textit{a good summary sentence is at least as informative as any single input sentence}, we develop novel schemes for path extraction from word graphs. Our contributions are as follows:  
\begin{enumerate}
    \item We present a novel scheme for path reranking in graph-based summarization. We show that, in practice, simple keyword counting performs better than complex baselines. For longer texts, we present an optional topic segmentation scheme. 
    \item We introduce a point-of-view (POV) conversion module to convert semi-extractive summaries into fully abstractive summaries. The new module by itself improves all scores on baseline methods, as well as our own. 
    \item Finally, We verify our model on datasets beyond those traditionally used in literature to provide a strong baseline for future research. 
\end{enumerate}

With just an off-the-shelf part-of-speech (POS) tagger and a list of stopwords, our model can be applied across different types of dialogue summarization. 

\begin{figure*}
  \centering
  \includegraphics[width=\textwidth]{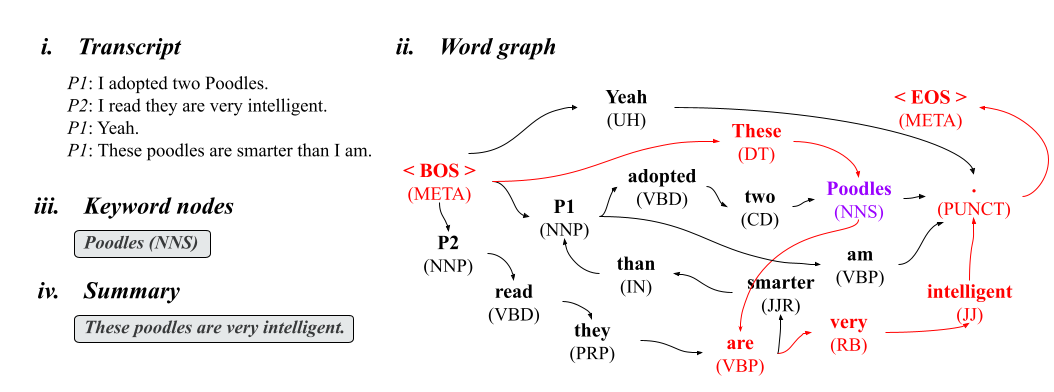}
  \caption{Construction of word graph. Red nodes and edges denote the selected summary path. Node highlighted in purple ("Poodles") is the only non-stopword node included in the $k$-core subgraph of the word graph. We use nodes from the $k$-core subgraph as keyword nodes. All original sentences from the unabridged input is present as a possible path from $v_{bos}$ to $v_{eos}$. Paths that contain more information than those original paths are extracted as summaries.}
  \label{fig:architecture}
\end{figure*}

\section{Background}

\subsection{Multi-sentence compression graphs}
Pioneered by \citet{filippova-2010-multi}, a Multi-Sentence Compression Graph (MSCG) is a graph whose nodes are words from the input text and edges are coocurrance statistics between adjacent words. During preprocessing, words “<bos>” (beginning-of-sentence) and “<eos>” (end-of-sentence) are prepended and appended, respectively, to every input sentence. Thus, all sentences from the input are represented in the graph as a single path from the \textit{<bos>} node ($v_{bos}$) to the \textit{<eos>} node ($v_{eos}$). Overlapping words among sentences will create intersecting paths within MSCG, creating new paths from $v_{bos}$ to $v_{eos}$, unseen in the original text. Capturing these possibly shorter but informative paths is the key to performant summarization with MSCGs. 

\citet{ganesan-etal-2010-opinosis} introduce an abstractive sentence generation method from word graphs to produce opinion summaries. \citet{tixier2016graph} show that nodes with maximal neighbors – a concept captured by graph degeneracy – likely belong to important keywords of the document. Shortest paths from $v_{bos}$ to $v_{eos}$ are scored according to how many keyword nodes they contain. Subsequently, a budget-maximization scheme is introduced to find the set of paths that maximizes the score sum within designated word count \citep{tixier2017combining}. We also adopt graph degeneracy to identify keyword nodes in MSCG.

\subsection{Unsupervised Abstractive Dialogue Summarization}
Aside from MSCGs, unsupervised dialogue summarization usually employ end-to-end neural architectures. \citet{Zhang_Zhang_Zaheer_Ahmed_2021} and \citet{Zou_Lin_Zhao_Kang_Jiang_Sun_Zhang_Huang_Liu_2021} utilize text variational autoencoders (VAEs) \cite{DBLP:journals/corr/KingmaW13, bowman-etal-2016-generating} to decode conditional or denoised abridgements. \citet{fu2021repsum} reformulate summary generation into a self-supervised task by equipping auxiliary objectives to 
the training architecture. Among end-to-end frameworks we only include \citet{fu2021repsum} as our baseline, because the brittle nature of training text VAEs, coupled with the lack of detail on data and parameters used to train the models, render VAE-based methods beyond reproducible. 

\section{Summarization strategy}
In following subsections we outline our proposed summarization process. 
\subsection{Word graph construction}
First, we assemble a word graph $G$ from the input text. We use a modified version of \citet{filippova-2010-multi}'s algorithm for graph construction: 

\begin{itemize}
    \item Let $SW$ be a set of stopwords and $T = s_0, s_1, ...$ be a sequence of sentences in the input text.
    \item Decompose all $s_i \in T$ into a sequence of POS-tagged words.
    \begin{multline}
    s_i = (``bos", ``meta"), (w_{i,0}, pos_{i,0}), ... , \\ 
    (w_{i,n-1}, pos_{i,n-1}), (``eos", ``meta")
    \end{multline}
\item For every $(w_{i,j}, pos_{i,j}) \in s_i$ such that $w_i \notin SW$ and $s_i \in T$, add a node $v$ in $G$. If a node $v'$ with the same lowercase word $w_{i,k}$ and tag $pos_{i,k}$ such that $j \neq k$ exists, pair $(w_{i,j}, pos_{i,j})$ with $v'$ instead of creating a new node. If multiple such matches exist, select the node with maximal overlapping context ($w_{i,j-1}$ and $w_{i,j+1}$).  
    \item Add stopword nodes -- $(w_{i,j}, pos_{i,j}) \in s_i$ such that $w_{i,j} \in SW$ and $s_i \in T$ -- to $G$ with the algorithm described above.
    \item For all $s_i \in T$, add a directed edge between node pairs that correspond to subsequent words. Edge weight $w$ between nodes $v_1$ and $v_2$ is calculated as follows: 
        \begin{equation}
            w' = \frac{freq(v_1) + freq(v_2)}{(\sum_{s_i \in T} diff(i, v_1, v_2))^{-1}}
        \end{equation}
         \begin{equation}
        w'' = freq(v_1) * freq(v_2) 
        \end{equation}
        \begin{equation}
        w = w' / w''
        \end{equation}
    $freq(v)$ is the number of words from original text mapped to node $v$. $diff(i, v_1, v_2)$ is the absolute difference in word positions of $v_1$ and $v_2$ within $s_i$:
    \begin{equation}
        diff(i, v_1, v_2) = |k - j|
    \end{equation}
    , where $w_{ij}$ and $w_{ik}$ are words in $s_i$ that correspond to nodes $v_1$ and $v_2$, respectively. 
    
    In edge weight calculation, $w'$ favors edges with strong cooccurrence, while $w''^{-1}$ favors edges with greater salience, as measured by word frequency.
\end{itemize}

It follows from above that only a single \textit{<bos>} node and a single \textit{<eos>} node will exist once the graph is completed.

\subsection{Keyword extraction}
The resulting graph from the previous step is a composition that captures syntactic importance. Traditional approaches utilize centrality measures to identify important nodes within word graphs \cite{mihalcea-tarau-2004-textrank, erkan2004lexrank}. In this work we use graph degeneracy to extract keyword nodes. In a k-degenerate word graph, words that belong to $k$-core nodes of the graph are considered to be keywords. We collect $KW$, a set of nodes belonging to the $k$-core subgraph. The $k$-core of a graph is the maximally degenerate subgraph, with minimum degree of at least $k$.

\subsection{Path threshold calculation} \label{3.3}
Once keyword nodes are identified, we score every path from $v_{bos}$ to $v_{eos}$ that corresponds to a sentence from the original text. Contrary to previous research into word-graph based summarization, we use a simple keyword coverage score for every path: 

\begin{equation}
    Score_i =  \frac{|V_i \cap KW|}{|KW|}
\end{equation}, where $V_i$ is the set of all nodes in path $p_i$, a representation of sentence $s_i \in T$, within the word graph.
We calculate the path threshold $t$, the mean score of all sentences in the original text. Later, when summaries are extracted from the word graph, candidates with path score less than $t$ are discarded. We also experimented with setting $t$ as the minimum or maximum of all original path scores, but such configurations yielded inferior summaries influenced by outlier path scores. 

Our path score function is reminiscent of the diversity reward function in \citet{shang2018unsupervised}. However, we use the function as a measure of \textit{coverage} instead of \textit{diversity}. More importantly, we utilize the score as means to extract a threshold based on all input sentences, which is significantly different from \citet{shang2018unsupervised}'s utilization of the function as a monotonically increasing scorer in submodularity maximization.

\begin{table*}
\centering
\begin{tabular}{c | c | c | c | c }
\hline
\textbf{Dataset} & \textbf{Domain} & \textbf{Test files} & \textbf{Dialogue length (chars)} & \textbf{Summary length (chars)}\\
\hline
AMI & Meeting & 20 & 22,499 (4,665 words) & 1,808 (292 words) \\
ICSI & Meeting & 6 & 42,484 (8,926 words) & 2,271 (371 words) \\
DialogSum & Day-to-day & 500 & 633 (125 words) & 115 (19 words) \\
SAMSum & Day-to-day & 819 & 414 (84 words) & 109 (20 words) \\
MediaSum & Interview & 10,000 & 8,718 (1,562 chars) & 335 (59 words) \\
SummScreen & Screenplay & 2,130 & 23,693 (5,642 words) & 1,795 (342 words) \\
ADS & Debate & 45 & 2918 (534 words) & 882 (150 words) \\\hline
\end{tabular}
\caption{Statistics for benchmark datasets. All character-level and word-level statistics are averaged over  the test set and rounded to the nearest whole number.}
\label{table:datasets}
\end{table*}

\subsection{Topic segmentation}

\begin{figure}[t]
  \centering
  \includegraphics[width=\linewidth]{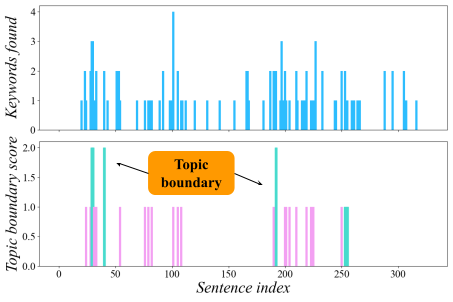}
  \caption{Topic segmentation on AMI meeting ID ES2005b. Green bars indicate sentence boundaries with highest topic distance.}
  \label{fig:example}
\end{figure}
For long texts, we apply an optional topic segmentation step. Our summarization algorithm is separately applied to each segmented text. Similar to path ranking in the next section, topics are determined according to keyword frequency. For every sentence in the input, we construct a \textit{topic coverage} vector $c$, a zero-initialized row-vector of length $|KW|$. Each column of the row vector is a binary representation signaling the presence of a single element in $KW$. Topic coverage vector of a path containing two keywords from $KW$, for instance, would contain two columns with 1.

Every transition between sentences is a potential topic boundary. Since each sentence (and corresponding path) has an associated topic coverage vector, we quantify the topic distance $d$ of a sentence with the next as the negative cosine distance of their topic vectors: 
\begin{equation}
    d_{i,i+1} = - \frac{c_i \cdot c_{i+1}}{\|c_i\| \|c_{i+1}\|}
\end{equation}

If $p$ is a hyperparameter representing the total number of topics, one can segment the original text at $p-1$ sentence boundaries with the greatest topic distance. Alternatively, sentence boundaries with topic distance greater than a designated threshold can be selected as topic boundaries. For simplicity, we proceed with the former segmentation setup (top-$p$ boundary) when necessary.

\subsection{Summary path extraction}
We generate a summary per-speaker. Our construction of the word graph allows fast extraction of sub-graphs containing only nodes pertaining to utterances from a single speaker. For each speaker subgraph, we generate summary sentences as follows: 

\begin{enumerate}
    \item We obtain $k$ shortest paths from $v_{bos}$ to $v_{eos}$ by applying the $k$-shortest paths algorithm \cite{yen1971finding} to our word graph.  
    \item Iterating from the shortest path, we collect any paths with keyword coverage score above the threshold calculated in \ref{3.3}.
    \item For each path found, we track the set of encountered keywords in $KW$. We stop our search if all keywords in $KW$ were encountered, or a pre-defined number of iterations (the search depth) is reached. 
\end{enumerate}

A good summary has to be both concise and informative. Intuitively, edge weights of the proposed word graph captures the former, while keyword thresholding prioritizes the latter. 

\begin{table*}
\centering
\begin{tabular}{l | c | c | c | c | c | c}
\hline
\multirow{2}{3em}{\textbf{Model}} & \multicolumn{3}{c|}{\textbf{AMI}} & \multicolumn{3}{c}{\textbf{ICSI}} \\
\cline{2-7}
& R1 & R2 & RL & R1 & R2 & RL \\ 
\hline
RepSum \citet{fu2021repsum} & 18.88 & 2.38 & 15.62 & - & - & - \\
\citet{filippova-2010-multi} & 33.47 & 6.21 & 15.15 & 26.53 & 3.69 & 12.09 \\
\citet{mehdad2013abstractive} & 34.62 & 6.49 & 15.41 & 27.20 & 3.57 & 12.55 \\
\citet{boudin2013keyphrase} & 34.21 & 6.37 & 14.92 & 26.90 & 3.64 & 12.18 \\
\citet{shang2018unsupervised} & 34.34 & 6.13 & 15.58 & 26.93 & 3.65 & 12.68 \\
\hline
\citet{filippova-2010-multi} $ _{+POV}$ & 34.16 & 6.35 & 15.27 & 26.79 & 3.81 & 12.21 \\
\citet{mehdad2013abstractive} $ _{+POV}$ & \textbf{35.39} & 6.59 & 15.54 & 27.48 & 3.65 & 12.66 \\
\citet{boudin2013keyphrase}  $ _{+POV}$ & 34.93 & 6.49 & 15.07 & 27.14 & 3.72 & 12.20 \\
\citet{shang2018unsupervised}  $ _{+POV}$ & 34.91 & 6.18 & \textbf{15.70} & 27.27 & 3.72 & \textbf{12.78} \\
\hline
Ours $ _{Pre Seg}$ & 32.21 & 5.55 & 14.85 & 27.60 & 4.43 & 11.66 \\
Ours $ _{Topic Seg}$ & 33.30 & 6.59 & 14.19 & 27.66 & 4.27 & 12.16 \\
Ours $_{Pre Seg + POV}$ & 33.66 & \textbf{6.85} & 14.17 & 27.80 & \textbf{4.56} & 11.77 \\
Ours $_{Topic Seg + POV}$ & 33.21 & 5.84 & 15.30 & \textbf{27.84} & 4.33 & 12.29 \\
\hline
\end{tabular}
\caption{Results on meeting summarization datasets. All reported scores are F-1 measures. Models with $POV$ indicate post-proceessing with our suggested POV conversion module. $PreSeg$ models utilize topic segmentations provided in \citet{shang2018unsupervised}, and $TopicSeg$ models intake unsegmented raw transcripts and perform the topic segmentation algorithm suggested in this paper. Results for RepSum are quoted from the original paper.}
\label{table:meetings}
\end{table*}

\subsection{POV conversion}
Finally, we convert our collected semi-extractive summaries into abstractive reported speech using a rule-based POV conversion module. We describe sentences extracted from our word graph as \textit{semi-extractive} rather than \textit{extractive}, to recognize the distinction between previously unseen sentences created from pieces of text, and sentences taken verbatim from the original text. Similar to existing \textit{extract-then-abstract} summarization pipelines \cite{mao2021dyle, liu2021combined}, our method hinges on the assumption that the extractive path-reranking step will optimize for \textit{summary content}, while the succeeding abstractive POV-conversion step will do so for \textit{summary style}. FewSum \cite{brazinskas-etal-2020-shot} also applies POV conversion in a few-shot summarization setting. FewSum conditions the summary generator to produce sentences in targeted styles, which is achieved by nudging the decoder to generate pronouns appropriate for each designated tone.

Popular literature has established that defining an all-encompassing set of rules for indirect speech conversion is infeasible \cite{Partee1973-PARTSA-18, li2011direct}. In fact, the English grammar is mostly descriptive rather than prescriptive -- no set of official rules dictated by a single governing authority exists. Even so, rule based POV conversion does provide a strong baseline compared to state-of-the-art techniques, such as end-to-end Transformer networks \cite{lee-etal-2020-converting}. In this study, we limit our scope to rule-based conversion because only the rule-based system among all tested methods in \citet{lee-etal-2020-converting} confers to the unsupervised nature of this paper. We encourage further research into integrating more advanced reported speech conversion techniques into the abstractive summarization pipeline. 

In this work, we apply four conversion rules:
\begin{enumerate}
    \item Change pronouns from first person to third person.
    \item Change modal verbs \textit{can}, \textit{may}, and \textit{must} to \textit{could}, \textit{might}, and \textit{had to}, respectively. 
    \item Convert questions into a pre-defined template: \textit{<Speaker> asks <utterance>}. 
    \item Fix subject-verb agreement after applying rules above.
\end{enumerate}
We notably omit prepend rules suggested in \cite{lee-etal-2020-converting}, because the input domain of our summarization system is unbounded, unlike with task-oriented spoken commands for virtual assistants. We also leave tense conversion for future research.  

\begin{table*}
\centering
\begin{tabular}{l | c | c | c | c | c | c}
\hline
\multirow{2}{3em}{\textbf{Dataset}} & \multicolumn{3}{c|}{\textbf{Our results}} & \multicolumn{3}{c}{\textbf{LEAD-3}} \\
\cline{2-7}
& R1 & R2 & RL & R1 & R2 & RL \\ 
\hline
DialogSum & \textbf{20.79} & 5.43 & 15.14 & 19.46 & \textbf{6.19} & \textbf{15.99} \\\hline
SAMSum & \textbf{26.48} & \textbf{9.69} & \textbf{19.65} & 21.93 & 8.52 & 18.65\\\hline
MediaSum & 7.19 & 1.79 & 5.66 & \textbf{8.58} & \textbf{3.19} & \textbf{6.62}\\\hline
SummScreen & \textbf{21.25} & \textbf{2.23} & \textbf{9.40} & 5.18 & 0.55 & 3.75\\\hline
ADS & \textbf{28.00} & \textbf{7.33} & \textbf{14.75} & 19.39 & 5.72 & 13.22\\
\hline
\end{tabular}
\caption{Results on day-to-day, interview, screenplay, and debate summarization datasets. All reported scores are F-1 measures. In our method, topic segmentation is applied to datasets with average transcription length greater than 5,000 characters (MediaSum, SummScreen), and POV conversion is applied to all datasets.}
\label{table:rest}
\end{table*}

\section{Experiments}

\subsection{Datasets}
We test our model on dialogue summarization datasets across multiple domains: 
\begin{enumerate}
    \item Meetings: \textit{AMI} \cite{mccowan2005ami}, \textit{ICSI} \cite{janin2003icsi}
    \item Day-to-day conversations: \textit{DialogSum} \cite{chen2021dialogsum}, \textit{SAMSum} \cite{gliwa-etal-2019-samsum}
    \item Interview: \textit{MediaSum} \cite{zhu-etal-2021-mediasum}
    \item Screenplay: \textit{SummScreen} \cite{chen2021summscreen}
    \item Debate: \textit{ADS} \cite{fabbri-etal-2021-convosumm}
\end{enumerate}

\autoref{table:datasets} provides detailed statistics and descriptions for each dataset. 

For AMI and ICSI, we conduct several ablation experiments with different components of our model omitted: semi-extractive summarization without POV conversion is compared with fully-abstractive summarization with POV conversion; utilization of pre-segmented text provided by \citet{shang2018unsupervised} is compared with application of topic segmentation suggested in this paper. 

\subsection{Baselines}
For meeting summaries, we compare our method with previous research on unsupervised dialogue summarization.
Along with \citet{filippova-2010-multi}, \citet{shang2018unsupervised}, and \citet{fu2021repsum}, we select \citet{boudin2013keyphrase} and \citet{mehdad2013abstractive} as our baselines. All but \citet{fu2021repsum} are word graph-based summarizers. 

For all other categories, we choose LEAD-3 as our unsupervised baseline. LEAD-3 selects the first three sentences of a document as the summary. Because summary distributions in several document types tend to be front-heavy \cite{grenander-etal-2019-countering, zhu-etal-2021-mediasum}, LEAD-3 provides a competitive extractive baseline with negligible computational burden.

\subsection{Evaluation}
We evaluate the quality of generated system summaries against reference summaries using standard ROUGE scores \cite{lin-2004-rouge}. Specifically, we use ROUGE-1 ($R1$), ROUGE-2 ($R2$), and ROUGE-L ($RL$) scores that respectively measure unigram, bigram, and longest common subsequence coverage.

\section{Results}

\subsection{Meeting summarization}
\autoref{table:meetings} records experimental results on AMI and ISCI datasets. In all categories, our method or a baseline augmented with our POV conversion module outperforms previous state-of-the-art.  

\subsubsection{Effect of suggested path reranking}
Our proposed path-reranking without POV conversion yields semi-extractive output summaries competitive with abstractive summarization baselines. Segmenting raw transcripts into topic groups with our method generally yields higher $F$-measures than using pre-segmented transcripts in semi-extractive summarization.

\subsubsection{Effect of topic segmentation} \label{5.1.2}
Summarizing pre-segmented dialogue transcripts results in higher $R2$, while applying our topic segmentation method results in higher $R1$ and $RL$. This observation is in line with our method's emphasis on keyword extraction, in contrast to keyphrase extraction seen in several baselines \cite{boudin2013keyphrase, shang2018unsupervised}. Models that preserve \textit{token adjacency} achieve higher $R2$, while models that preserve \textit{token presence} achieve higher $R1$. $RL$ additionally penalizes for wrong token order, but token order in extracted summaries tend to be well-preserved in word graph-based summarization schemes.

\subsubsection{Effect of POV conversion module}
Our POV conversion module improves benchmark scores on all tested baselines, as well as on our own system. It is only natural that a conversion module that translates text from semi-extractive to abstractive will raise scores on abstractive benchmarks. However, applying our POV module to \textit{already abstractive} summarization systems resulted in higher scores in all cases. We attribute this to the fact that previous abstractive summarization systems do not generate sufficiently reportive summaries; past research either emphasize other linguistic aspects like hyponym conversion \cite{shang2018unsupervised}, or treat POV conversion as a byproduct of an end-to-end summarization pipeline \cite{fu2021repsum}.

\subsection{Day-to-day, interview, screenplay, and debate summarization}
\begin{figure}[t]
  \centering
  \includegraphics[width=\linewidth]{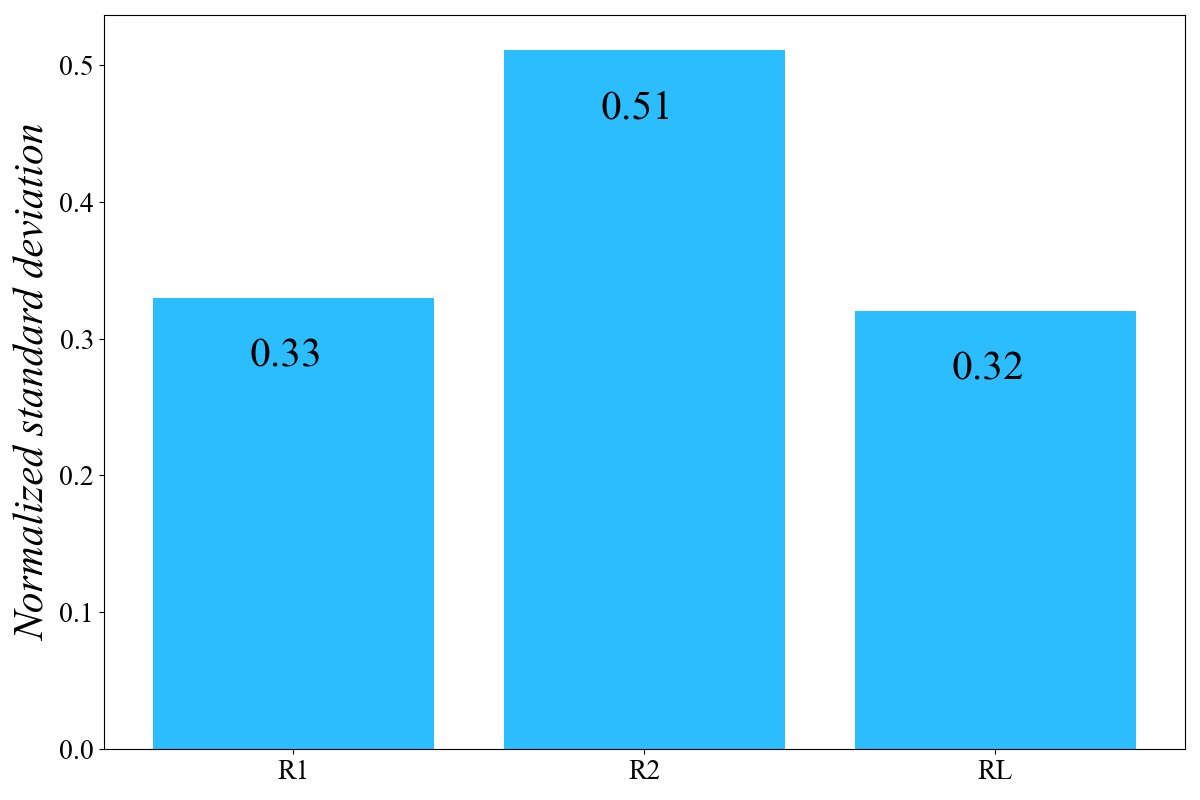}
  \caption{Normalized standard deviation (also called coefficient of variance) of R1, R2, and RL scores across all datasets. Normalized standard deviation is calculated as $\sigma/\Bar{x}$, where $\sigma$ is the standard deviation and $\Bar{x}$ is the mean.}
  \label{fig:normalized_std}
\end{figure}

Our method outperforms the LEAD-3 baseline on most benchmarks (\autoref{table:rest}). The model shows consistent performance across multiple domains in $R1$ and $RL$, but shows greater inconsistency in $R2$. Variance in the latter metric can be attributed, as in \ref{5.1.2}, to our model's tendency to optimize for single keywords rather than keyphrases. Robustness of our model, as measured by consistency of ROUGE measures across multiple datasets, is shown in \autoref{fig:normalized_std}.

\begin{table*}
\centering
\begin{tabular}{c | c }
\textbf{Transcript} & \textbf{Summary} \\
\hline
\makecell[l]{
    \textit{Maya: Bring home the clothes that are hanging outside} \\
    \textit{Maya: All of them should be dry already and} \\
    \textit{\qquad \quad it looks like it's going to rain} \\
    \textit{Boris: I'm not home right now} \\
    \textit{Boris: I'll tell Brian to take care of that}\\
    \textit{Maya: Fine, thanks} \\
} &
\makecell[l]{
     \textit{bring home the clothes that are hanging outside} \\
     \textit{boris 'll tell brian to take care of that}
} \\
\hline
\multicolumn{2}{l}{\textit{Keywords: 'care', 'clothes', 'home', 'thanks'}} \\
\hline \hline \hline

\makecell[l]{
    \textit{Megan: Are we going to take a taxi to the opera?} \\ 
    \textit{Joseph: No, I'll take my car.} \\ 
    \textit{Megan: Great, more convenient}
} &
\makecell[l]{
    \textit{are we going to take a taxi to the opera ?} \\ 
    \textit{no , joseph 'll take my car}
} \\
\hline
\multicolumn{2}{l}{\textit{Keywords: 'car', 'convenient', 'taxi', 'opera'}} \\
\hline \hline \hline

\makecell[l]{
    \textit{Anne: You were right, he was lying to me :/} \\ 
    \textit{Irene: Oh no, what happened?} \\ 
    \textit{Jane: who?} \\
    \textit{Jane: that Mark guy?} \\
    \textit{Anne: yeah, he told me he's 30, today I saw his} \\
    \textit{\qquad \quad passport - he's 40} \\
    \textit{Irene: You sure it's so important?} \\
    \textit{Anne: he lied to me Irene}
} &
\makecell[l]{
    \textit{he lied to me he 's 30 , today anne saw his} \\ 
    \textit{passport - he 's 40 yeah , he told me} \\
    \textit{ oh no , what happened? who ? } \\
    \textit{annerene he lied to me : /}
} \\
\hline
\multicolumn{2}{l}{\textit{Keywords: 'guy', '/', 'passport', 'yeah', 'today'}} \\
\hline \hline \hline

\end{tabular}
\caption{Summarizing the SAMSum corpus \cite{gliwa-etal-2019-samsum}.}
\label{table:examples}
\end{table*}

Notably, our method falters in the MediaSum benchmark. Compared to other benchmarks, MediaSum's reference summaries display heavy positional bias towards the beginning of its transcripts, which benefits the LEAD-3 approach. It also is the only dataset in which references summaries are not generated for the purpose of summary evaluation, but are scraped from source news providers. Reference summaries for MediaSum utilize less reported speech compared to other datasets, and thus our POV module fails to boost the precision of summaries generated by our model.

\section{Conclusion}
\subsection{Improving MSCG summarization}
This paper improves upon previous work on multi-sentence compression graphs for summarization. We find that simpler and more adaptive path reranking schemes can boost summarization quality. We also demonstrate a promising possibility for integrating point-of-view conversion into summarization pipelines. 

Compared to previous research, our model is still insufficient in keyphrase or bigram preservation. This phenomenon is captured by inconsistent $R2$ scores across benchmarks. We believe incorporating findings from keyphrase-based summarizers \cite{riedhammer2010long, boudin2013keyphrase} can mitigate such shortcomings. 

\subsection{Avenues for future research}
While our methods demonstrate improved benchmark results, its mostly heuristic nature leaves much room for enhancement through integration of statistical models. POV conversion in particular can benefit from deep learning-based approaches \cite{lee-etal-2020-converting}. With recent advances in unsupervised sequence to sequence transduction \cite{li2020_Optimus, He2020A}, we expect further research into more advanced POV conversion techniques will improve unsupervised dialogue summarization.    

Another possibility to augment our research with deep learning is through employing graph networks \cite{cui-etal-2020-enhancing} for representing MSCGs. With graph networks, each word node and edge can be represented as a contextualized vector. Such schemes will enable a more flexible and interpolatable manipulation of syntax captured by traditional word graphs.

One notable shortcoming of our system is the generation of summaries that lack grammatical coherence or fluency (\autoref{table:examples}). We intentionally leave out complex path filters that gauge linguistic validity or factual correctness. We only minimally inspect our summaries to check for inclusion of verb nodes, as in \citet{filippova-2010-multi}. Our system can be easily augmented with such additional filters, which we leave for future work.

\bibliography{anthology,custom}
\bibliographystyle{acl_natbib}

\end{document}